# A REINFORCED EVOLUTION-BASED APPROACH TO MULTI-RESOURCE LOAD BALANCING

[1]Leszek Sliwko

[1]Institute of Applied Informatics, Wrocław University of Technology, Wrocław, Poland

E-mail: lsliwko@gmail.com

## ABSTRACT

This paper presents a reinforced genetic approach to a defined *d-resource system optimization problem*. The classical evolution schema was ineffective due to a very strict feasibility function in the studied problem. Hence, the presented strategy has introduced several modifications and adaptations to standard genetic routines, e.g.: a migration operator which is an analogy to the biological random genetic drift.

## 1. INTRODUCTION

A fundamental goal in computer science is to provide an algorithm which would determine an optimal solution in acceptable time. Computational Complexity Theory is the field which studies the efficiency of computation; its major goals are to find efficient algorithms for natural problems or to show that no efficient solutions exist.

NP-hard (Nondeterministic Polynomial-time hard), represents a class of problems which are 'at least as difficult as problems in NP' [7][19]. NP-complete problems can be solved by means of exhaustive search. However the time to wait for the solution grows unacceptably with the problem size as the number of iterations needed to solve the problem becomes tremendous [19]. In such a case the best we can hope for are super-polynomial time algorithms. The 'P versus NP' problem is one of the seven open Millennium Prize Problems of the Clay Mathematics Institute. It is now commonly believed that $P \neq NP$ [7][19][22] and it is rather unlikely that there can ever be any efficient (Polynomial Time) exact algorithms able to solve NP-hard problems.

NP-hard problems may be of any type, ranging from search, decision, or optimization problems to feasibility problems [19]. Discrete optimization problems are generally NP-hard problems.

In Computational Complexity Theory, a meta-heuristic algorithm (the term 'meta-heuristic' is originally derived from the Greek 'μετά' (*a higher level*) and 'ευρισκειν' (*to discover*)) is a scientific method that solves a problem with the help of iterative stochastic processes. A heuristic algorithm usually gives up the optimality of the solution in order to finish within a satisfactory timeframe [22]. Generally speaking, it is able to find quite a good solution, but there is no proof that the result could not be better or that the solution found by the heuristic algorithm would be feasible in the first place.

Genetic Algorithms play a significant role among meta-heuristics schemas due to their universality and scalability [8][9]. Additionally, Genetic Algorithms are able to fulfil their tasks even in the presence of noise [15] and, unlike other AI schemas; they do not break even if inputs change significantly [4]. Genetic Algorithms may also offer significant benefits over a more typical search of optimization techniques as they can be applied to several problems without any major changes in their design [2][11][14][20].

In this paper, a reinforced evolution approach to defined NP-hard optimization problem is presented. The studied problem is quite unusual as it introduces a strict feasibility function. Several interesting adaptations and modifications to the classical evolution schema were required in order to create a satisfying strategy, i.a. authors developed original *migration* operator which emulates biological random genetic drift [6].

The proposed schema has been implemented and the experiment results as well as the conclusions are presented here.

## 2. PROBLEM FORMULATION

**Definition 1.** Let us define $\Lambda = (\tau, \eta, \psi, a, r, c)$ as problem space and our system as twice $(\Lambda, \mu)$.

In the ***d-resource system optimization problem***, we receive a set $\tau$ of $l$ mobile tasks: $\tau = \{t_1, t_2, ..., t_l\}$ and a set $\eta$ of $m$ fixed nodes: $\eta = \{n_1, n_2, ..., n_m\}$. We call $\mu : \tau \to \eta$ a *task assignment* function, i.e.: every task has to be assigned to the node.

We also consider:
- $\psi = \{i_1, i_2, ..., i_d\}$ as a set of all different kinds of resources, e.g.: for $d = 3$ we could define $\psi = \{CPU, memory, network\}$.
- $a : \psi \times \eta \to \mathbb{N} \cup \{0\}$ as fixed *available resources* on the nodes, i.e.: $a_i(n)$ is the available level (integer value) of a resource $i$ on the node $n$.
- $r : \psi \times \tau \to \mathbb{N} \cup \{0\}$ as fixed *required resources* for tasks, i.e.: $r_i(t)$ is the required level (integer value) of a resource $i$ of a task $t$.
- $c : \tau \to \mathbb{N} \cup \{0\}$ as a *task migration cost* function, i.a.: $c(t)$ can mean the amount of hours a developer has to spend deploying task $t$ on the node.

For every node $n \in \eta$, we define a set $A_n = \{t \in \tau : \mu(t) = n\}$ of all tasks assigned to the node $n$. We consider the system $(\Lambda, \mu)$ as *stable* iff:

$$\sum_{t \in A_n} r_i(t) \le a_i(n), \text{ for every } n \in \eta, i \in \psi \quad (1)$$

Otherwise, the system $(\Lambda, \mu)$ is *overloaded*.

Each task $t$ is initially assigned by the *task assignment* function $\mu_0$ to a node $n$; during *system transformation* $(\mu_0 \to \mu_1)$, the task $t \in \tau$ can be reassigned to any other node $n \in \eta$. The process of moving the task to another node is called here *task migration* and it generates *task reassigning cost*:

$$c_{(\mu_0 \to \mu_1)}(t) = \begin{cases} 0, & \mu_0(t) = \mu_1(t) \\ c(t), & \mu_0(t) \ne \mu_1(t) \end{cases}$$

Every *system transformation* process $(\mu_0 \to \mu_1)$ has its *system transformation cost*:

$$c_{(\mu_0 \to \mu_1)} = \sum_{t \in \tau} c_{(\mu_0 \to \mu_1)}(t)$$

Consider an initial *task assignment* $\mu_0$; the *task assignment* $\mu^*$ is optimal for $\mu_0$, iff $\mu^*$ renders system $(\Lambda, \mu^*)$ *stable* and:

$$c_{(\mu_0 \to \mu^*)} \le c_{(\mu_0 \to \mu)}, \text{ for every } stable \text{ system } (\Lambda, \mu) \quad (2)$$

N.b.: when $(\Lambda, \mu_0)$ is *stable* for the initial *task assignment* $\mu_0$, the *system transformation cost* equals 0 as it is considered optimal.

## 3. WORKING EVOLUTION SCHEMA

Nowadays, stochastic algorithms are used ever more frequently, ranging from designing a concert hall with optimal acoustic properties [18], evolving wire antennas [1], predicting the future performance of stocks [12] to even guessing the location of earthquake hypocenters [17].

The classical approach defines a binary vector as genotype mapping [20]. However, in our approach we found it more suitable to denote the search space by $\Phi_g = \eta^{|\tau|}$, therefore defining the $|\eta|$-ary language as a candidate solution representation. Each genotype $G_\alpha$ within search space $\Phi_g$ represents a *system transformation* $(\mu_0 \to \mu_\alpha)$ and is defined as an arbitrary vector:

$$G_\alpha = (\mu_\alpha(t_1), \mu_\alpha(t_2), ..., \mu_\alpha(t_l)) \in \Phi_g$$

We adapt the *system transformation cost* in the fitness function equation $f_{fit} : G \to \mathbb{N} \cup \{0\}$. For genotype $G_\alpha$ it is defined:

$$f_{fit}(G_\alpha) = \sum_{t \in \tau} c(t) - c_{(\mu_0 \to \mu_\alpha)}$$

In our experiment, usually less than 4% of all the possible genotypes represented a feasible solution (i.e. solution shall satisfy (1) and (2)) at all, not even mentioning its optimality. It can be also noticed that feasible solutions were usually scattered all over the problem search space.

To confront this optimization enigma, we utilized a number of standard genetic operators [4][8][9][15] as well as a few new ones which had been developed for the purpose of this experiment.

We did not follow the classical schema of forming a new population with the current one as a base in every iteration of the algorithm [8][15]. Instead, we enabled our individuals to recombine freely, even with the individuals from the previous population if suitable.

Our evolution model operates in iterations. One system cycle is presented below:

1. *Selection* for *crossover* (25% of individuals are chosen)
2. *Selection* for *mutation* (5% of individuals are chosen)
3. *Termination* of *unstable* genotypes (with a 10% survival rate)
4. *Termination* of the weakest individuals (up to 20% of the population size)
5. *Migration* (filling up the gene pool to the population limit)

Since the initial generation, the size of the population will never decrease below a defined number, called here the *living space*. We experimented with various population sizes and we found out that the best results are obtained for the sizes between:

$$|\tau\|\eta| \leq X \leq 2|\tau\|\eta|$$

For the sizes less than $|\tau\|\eta|$, the population sometimes lacked genotype diversity and the algorithm could not escape from local minima. The sizes above $2|\tau\|\eta|$ generally provided no significant gain in the algorithm performance, so we decided to use this value as our default.

Several genetic operators were adapted to serve our evolution model needs:

### 3.1 Selection

We have employed *Tournament Selection* algorithm [5]. It is one of the selection methods in Genetic Algorithms which runs a 'tournament' among a few individuals randomly chosen from the population. The individuals are paired and they perform a 'match' where the better one is selected for the next 'turn'. The tournament is played in a repetitive manner with the individual winners becoming participants again. The winner of the last pair is then selected for a recombination.

*Tournament Selection* allows the selection pressure to be easily adjusted by changing the tournament size. If the tournament size is larger, weak individuals have a smaller chance to be selected.

### 3.2 Crossover

In Genetic Algorithms, *crossover* is analogous to a reproduction and biological *crossover* [6], upon which genetic algorithms are based. The operator attempts to combine some elements of the existing solutions in order to create a new solution comprising certain features of each of the parents.

The combination of parents' chromosomes is usually made by selecting one or more *crossover* points randomly, splitting input chromosomes onto the selected points and then linking those sequences of different chromosomes to eventually compose new genotypes [5]. For a detailed survey of crossover techniques please refer to [15][16].

In our approach, however, we produce only one offspring. We randomly copy one part of a chromosome from the first parent and complete it with some genetic material from the other parent (one-point crossover [2][20]).

### 3.3 Mutation

Optimization algorithms are sometimes prone to local minima – a point where the fitness function value is bigger than its neighbours, but possibly smaller than at some distant point in the search space. The genetic algorithm can overcome this deficiency with the *mutation* operator, forcing the genetic diversity, subject to right settings.

In our case, the *mutation* operator does not alter the structure of the target genotype. Instead, it clones the whole structure and changes one random gene to a random value within possible gene space. This feature proved to be capable of preventing the slowdown of the evolution process [11].

### 3.4 Termination

The *termination* operator plays a very important role of eliminating weak individuals from the population and thus improving the whole solution pool and producing *living space* for the coming individuals. On the other hand, the size of the population has to be limited in order to keep a good computing and also due to computer limitations.

### 3.5 Migration

In population genetics, a random genetic drift (also known as a gene flow) is a transfer of alleles of genes from one population to another [6]. The immigration takes place strictly by chance and it may result in adding some new genetic material to the established gene pool of the current population.

The main purpose of introducing the *migration* operator was the necessity to compensate for the intolerant elimination of individuals from the population. Therefore, in our case, *migration*, along with the *mutation* operator, are main sources of genotype diversity in the population.

In the *migration* phase, we randomly create a genotype. Only *stable* genotypes are accepted, thus in some setups it takes a number of draws to create a valid individual.

It is a very crucial process. In our model, at least 20% of the population is eliminated in iteration. The random genetic drift feature helps the system to maintain a constant number of individuals in the population.

## 4. EXPERIMENT

Similar to our previous system [21], the simulation was performed with the help of JMASB (Java Multi-Agent System Balancer). The framework was initially developed for agent-based system performance analysis, and it enables the researcher to test even complex schemes when planning a resource management strategy. It was quite a challenge itself to recode the framework so that we could utilize the evolution approach.

The system is designed to solve **d-resource system optimization problem** on average machine and so an accessible configuration has a considerable impact on the system performance. Thus, we decide to use low-end hardware to process, which should be widely available in our market:

| Testing environment | |
|---|---|
| Processor | Intel Pentium IV (Northwood) 2000MHz |
| Memoey | 512MB PC2700 DDR SDRAM Kingston |
| Main board | Intel S845WD1-E (Intel 845E) |
| Java Virtual Machine | Java(TM) 2 Runtime Environment, Standard Edition 1.4.2_15 |
| Disc array | FastTrack100 RAID0 (Stripe 2+0) |
| Hard drive (s) | 2 x Samsung SP1614N 160GB |
| System | Windows 2000 Professional SP4 |

**Figure 1.** Testing environment

Our earlier work [21] demonstrated that a non-deterministic strategy is able to handle a potentially unlimited number of resources; in this simulation we used various resources: CPU, memory and network. As we have already tested the system functionalities and correctness, we will focus on testing general usefulness of the presented approach to discrete optimization problems.

### 4.1 Research principles

For the purpose of the experiment, five different strategies: FULLSCAN, GREEDY, BALANCE, GENETIC and EVOLVE, were deployed; the goal was to compare the system with other common optimization schemas.

The FULLSCAN strategy, as the name suggests, performs a full search over all available configurations. The FULLSCAN strategy guarantees a globally optimal solution under appropriate modelling assumptions. It cannot be considered an efficient strategy due to a large scale of computation level required – in bigger instances of a problem we could not wait for this algorithm to finish and so we decided to terminate it if the continuous computation took more than a week.

GREEDY is an algorithm that follows solving problems of making locally optimum choices in its every iteration. This is a simple but generally effective strategy. A brief description of this approach can be found in [3].

The BALANCE strategy is an adaptation of Google AdWords schema [13]. The original Google strategy is based on the remaining budgets comparison. We noticed there an analogy to our nodes capacity. In our experiment its results were generally better than the GREEDY algorithm.

The GENETIC strategy is an implementation of a original Holland's Genetic Algorithm approach [8][9]. The results from this strategy are used as referring point for the result of experiments.

The EVOLVE strategy is basically what we are describing in this paper. Among all the defined strategies, EVOLVE and GENETIC are not deterministic, thus we run them up to forty times and compare the obtained results (Fig. 4).

### 4.2 Experiment configuration

The initial configuration consists of eight nodes with assigned levels of available resources (Fig. 2):

| Node name | Available resources | | |
|---|---|---|---|
| | CPU | Memory | Network |
| Node1 | 60 | 60 | 50 |
| Node2 | 70 | 40 | 50 |
| Node3 | 70 | 70 | 70 |
| Node4 | 80 | 50 | 90 |
| Node5 | 60 | 80 | 50 |
| Node6 | 60 | 70 | 50 |
| Node7 | 80 | 70 | 80 |
| Node8 | 80 | 90 | 60 |

**Figure 2.** Experiment nodes available resources

Forty jobs were generated (Fig. 3). We also defined the *task migration cost* for each of them:

| Task name | Needed resources | | | Task migration cost |
|---|---|---|---|---|
| | CPU | Memory | Network | |
| J01 | 7 | 15 | 7 | 4 |
| J02 | 20 | 3 | 16 | 5 |
| J03 | 1 | 1 | 1 | 4 |
| J04 | 18 | 13 | 9 | 7 |
| J05 | 14 | 10 | 1 | 10 |
| J06 | 3 | 12 | 13 | 3 |
| J07 | 11 | 18 | 12 | 6 |
| J08 | 1 | 4 | 8 | 6 |
| J09 | 4 | 3 | 17 | 4 |
| J10 | 8 | 19 | 19 | 4 |
| J11 | 5 | 9 | 18 | 8 |
| J12 | 16 | 14 | 3 | 6 |
| J13 | 6 | 5 | 17 | 4 |
| J14 | 18 | 11 | 13 | 5 |
| J15 | 10 | 9 | 12 | 1 |
| J16 | 12 | 17 | 14 | 9 |
| J17 | 3 | 6 | 8 | 5 |
| J18 | 8 | 12 | 3 | 5 |
| J19 | 15 | 12 | 8 | 7 |
| J20 | 4 | 8 | 6 | 1 |
| J21 | 12 | 10 | 5 | 7 |
| J22 | 3 | 19 | 16 | 2 |
| J23 | 6 | 19 | 1 | 5 |
| J24 | 19 | 11 | 2 | 3 |
| J25 | 14 | 8 | 15 | 4 |
| J26 | 4 | 15 | 7 | 10 |
| J27 | 20 | 19 | 5 | 2 |
| J28 | 16 | 2 | 3 | 8 |
| J29 | 16 | 10 | 3 | 6 |
| J30 | 1 | 1 | 3 | 5 |
| J31 | 19 | 18 | 1 | 4 |
| J32 | 6 | 14 | 3 | 10 |
| J33 | 3 | 10 | 3 | 2 |
| J34 | 2 | 8 | 1 | 4 |
| J35 | 8 | 9 | 9 | 9 |
| J36 | 8 | 15 | 13 | 4 |
| J37 | 19 | 8 | 5 | 2 |
| J38 | 16 | 20 | 1 | 2 |
| J39 | 15 | 20 | 4 | 8 |
| J40 | 6 | 13 | 10 | 10 |

**Figure 3.** Experiment tasks and their migration costs

In the course of the experiment, several different scenarios were tested. There we present three random samples together with their output (for the reader's convenience the initially overloaded nodes have been **bolded**):

| Initial configuration | |
|---|---|
| Node name | Tasks |
| Node01 | J01, J04, J14, J16 |
| **Node02** | J08, J11, J12, J15, J18 |
| **Node03** | J02, J03, J06, J07, J13, J19, J20 |
| Node04 | J05, J09, J10, J17 |
| Node05 | (node not available during test) |
| Node06 | (node not available during test) |
| Node07 | (node not available during test) |
| Node08 | (node not available during test) |
| Experiment results | |
| Strategy name | System transformation cost |
| Highest possible migration cost (all tasks migrated) | 104 |
| FULLSCAN strategy (optimal cost) | 5 *(stategy needed less than 2 minutes)* |
| EVOLVE strategy | 11 (median) / 12.95 (average) |
| GENETIC strategy | 11 (median) / 13.23 (average) |
| GREEDY strategy | 24 |
| BALANCE strategy | 18 |

**Figure 4.** The first test

All the strategies finished their tasks (Fig. 4) in a relatively short time. Both GREEDY and BALANCE ended almost immediately rendering pretty good results. An optimal solution was computed in less than 2 minutes by the FULLSCAN algorithm. We set a 30 second time limit for GENETIC and EVOLVE strategies. Both approaches behaved similar and an optimal *system transformation cost* was found in just few cycles of their runs.

| Initial configuration | |
|---|---|
| Node name | Tasks |
| **Node01** | J01, J03, J04, J06, J16, J20, J26 |
| **Node02** | J08, J09, J17, J18, J25, J28, J30 |
| Node03 | J07, J14, J19, J22, J29 |
| **Node04** | J10, J11, J23, J24, J27 |
| Node05 | J02, J13 |
| Node06 | J05, J12, J15, J21 |
| Node07 | (node not available during test) |
| Node08 | (node not available during test) |
| Experiment results | |
| Strategy name | System transformation cost |
| Highest possible migration cost (all tasks migrated) | 156 |
| FULLSCAN strategy (optimal cost) | 15 *(stategy needed 53 hours to complete)* |
| EVOLVE strategy | 19 (median) / 21.90 (average) |
| GENETIC strategy | 20 (median) / 23.48 (average) |
| GREEDY strategy | 49 |
| BALANCE strategy | 32 |

**Figure 5.** The second test

The second test (Fig. 5) demonstrated that we could not use the exhaustive scan approach for bigger instances of a problem. We were able to compute an optimal solution with this approach, but it took as many as two days.

Again, GREEDY and BALANCE finished in a few seconds. Strangely, the GREEDY outcome is worse than in the previous test, but this is how deterministic algorithms sometimes work.

The EVOLVE strategy computed solutions with the *system transformation cost* equal to 21.90 on average. The results from GENETIC schema were slightly worse, averaging on 23.48. The timeouts for both strategies were set to 10 minutes for each run; the EVOLVE strategy found the optimal solution cost in one of its runs.

| Initial configuration | |
|---|---|
| Node name | Tasks |
| Node01 | J01, J04, J16 |
| Node02 | J11, J18, J27, J28 |
| Node03 | J02, J06, J07, J19 |
| Node04 | J05, J09, J10, J17, J24, J39 |
| Node05 | J14, J31, J34, J35, J38 |
| Node06 | J08, J12, J15, J21, J25, J30, J33, J36 |
| Node07 | J03, J13, J20, J22, J26, J29, J37, J40 |
| Node08 | J23, J32 |
| Experiment results | |
| Strategy name | System transformation cost |
| Highest possible migration cost (all tasks migrated) | 211 |
| FULLSCAN strategy (optimal cost) | (strategy could not finish its task within 7 days) |
| EVOLVE strategy | 26 (median) / 26.75 (average) |
| GENETIC strategy | 31.5 (median) / 31.33 (average) |
| GREEDY strategy | 39 |
| BALANCE strategy | 33 |

**Figure 6.** The third test

The last test (Fig. 6) demonstrated that the evolution schema outperforms deterministic strategies in terms of correctness and scalability. The timeout for this simulation was set to 60 minutes and during that time EVOLVE found the lowest *system transformation cost* equal to 19. We were not able to compute optimal solution, due to long time the computation would require (over $10^{36}$ iterations are needed at the worst), but we believe this value is close to optimal.

### 4.3 Outcome analysis

Working solutions were found for all the strategies. The results of our experiment are presented as statistics in table below (Fig. 7):

| Problem | | Experiment results | | | | | | |
|---|---|---|---|---|---|---|---|---|
| Instance size | Search space size | Highest possible migration cost | FULLSCAN strategy (optimal cost) | EVOLVE strategy | | | | |
| | | | | Med | Avg | Min | Max | StDev | Count |
| 20 jobs (4 nodes) | 1.10E+12 | 104 | 5 | 11 | 12.95 | 5 | 26 | 6.37 | 40 |
| 25 jobs (5 nodes) | 2.98E+17 | 125 | 14 | 18 | 21.60 | 16 | 67 | 9.13 | 40 |
| 30 jobs (6 nodes) | 2.21E+23 | 156 | 15 | 19 | 21.90 | 15 | 45 | 7.80 | 40 |
| 35 jobs (7 nodes) | 3.79E+29 | 185 | -- | 24 | 26.45 | 22 | 57 | 7.05 | 40 |
| 40 jobs (8 nodes) | 1.33E+36 | 211 | -- | 26 | 26.75 | 19 | 49 | 7.54 | 40 |
| Problem | | Experiment results | | | | | | |
| Instance size | Search space size | GREEDY strategy | BALANCE strategy | GENETIC strategy | | | | |
| | | | | Med | Avg | Min | Max | StDev | Count |
| 20 jobs (4 nodes) | 1.10E+12 | 24 | 18 | 11 | 13.23 | 5 | 22 | 5.86 | 40 |
| 25 jobs (5 nodes) | 2.98E+17 | 18 | 31 | 20 | 22.35 | 16 | 48 | 7.75 | 40 |
| 30 jobs (6 nodes) | 2.21E+23 | 49 | 32 | 20 | 23.43 | 18 | 48 | 7.20 | 40 |
| 35 jobs (7 nodes) | 3.79E+29 | 28 | 29 | 27 | 30.83 | 24 | 67 | 10.88 | 40 |
| 40 jobs (8 nodes) | 1.33E+36 | 39 | 33 | 31.5 | 31.33 | 20 | 81 | 11.32 | 40 |

**Figure 7.** Experiment results

The EVOLVE strategy has high estimated standard deviation (*StDev*) to average (*Avg*) migration cost ratio. In forty runs of each instance of the problem, the algorithm usually found two to three solutions with extreme migration cost. However, most of the given solutions were of a good quality, which resulted in small differences between average (*Avg*) and median (*Med*) of the migration costs. In all cases the best migration cost (*Min*) found was better that results of GREEDY and BALANCE strategies. In two cases we were also able to confirm optimality of the solutions.

The FULLSCAN strategy was characterized by a long computation time. Its outcome is always a global optimum, however, the trade-off of this schema is the necessity to test a tremendous number of candidate solutions. Still, this approach is pretty useful in smaller instances of a given problem (up to thirty jobs and six nodes).

The GREEDY strategy performed the worst, which came as a surprise. It seems not well suited for this kind of combinatorial optimization problems where solution space swarms with local minima.

The BALANCE strategy behaved well. The Google AdWords idea of utilizing the remaining capacity comparisons finds its application here. It is a very fast strategy which generally renders good results.

The biggest drawback of the EVOLVE and GENETIC schemas is their complexity. The Genetic Algorithm schema is universal and can be easily adapted to various problems. However, this has its cost in a highly abstract model it provides. Such models usually require more redundancy in the computer code of their implementations. It has also its execution time demand. The GENETIC strategy has been slightly faster and computed on average 25% cycles more in defined time. The EVOLVE schema spent most of the algorithm execution time (over 70%) on the *migration* step, due to a large number of random draws that had to be performed in order to create an acceptable individual.

Particularly, computing the first population takes a lot of time as the algorithm has to fill up the whole genotype pool which is initially empty. After this step, migration adds no more than 20% of the maximum population size, usually ~1-3% in one algorithm iteration, hence it is bearable.

It could be discussed if we should not have computed a table of sample solutions first, and then tried to randomize them. Our research suggests that we could have possibly distributed the produced individuals in certain areas of the problem search space. This technique, called seeding, can scientifically improve the algorithm outcome, as it was referred to in [10].

Nevertheless, our strategy is really remarkable in larger instances of the problem regarding both completeness and scalability. The results are well refined, even in comparison to the exhaustive search approach.

## 5. CONCLUSION

In this paper, we have demonstrated the usefulness of evolution schemas to a combinatorial optimization problem of the presented proprieties. The defined problem was characterized by a rigorous feasibility requirement which caused the solution space to be chaotic, rendering the classic Genetic Algorithms approach less effective. Therefore, the presented strategy introduced several modifications and adaptations to standard genetic routines.

In the course of the research, we defined a new genetic operator called *migration*. The *migration* operator is analogous to the biological random genetic drift. Its main purpose was to compensate for a very rigid elimination of individuals of the population.

The presented strategies were also tested against optimal solutions computed with the help of exhaustive search. This experiment demonstrated that the algorithm met its requirements and the evolution strategy proved competitive enough. In terms of correctness and scalability our strategy performs well. It is necessary, however, to mention the drawbacks of our schema.

First, the process spends over 70% of execution time on the *migration* step. It has been already mentioned that the introduction of sample solution tables would possibly reduce the computation power required to perform the algorithm iteration. Additionally, we could possibly distribute the produced individuals in certain areas of problem search space.

Another drawback of the presented approach can be the number of solution feasibility checks required to probe a genotype. Our current model has to iterate through every node and its tasks to check if (1) is satisfied. We could probably group the genotypes by their similarity and then compute checks exclusively on the varying nodes.

The experiment results were more than satisfactory. However, in order to address performance issues we are considering the introduction of the seeding schemas and various other model optimizations in our future work on the system.

## 6. REFERENCES

[1] Altshuler, E.E. and Linden, D.S., "Design of a wire antenna using a genetic algorithm", *Journal of Electronic Defence*, Vol.20, No.7, 1997, pp.50-52.

[2] Back, T., Hammel, U. and Schwefel, H-P., "Evolutionary computation: comments on the history and current state", *IEEE Transactions on Evolutionary Computation*, Vol.1, Iss.1, 1997, pp.3-17.

[3] Becker, A. and Geiger, D., "Optimization of Pearls method of conditioning and greedy-like approximation algorithms for the vertex feedback set problem", *Artificial Intelligence*, Vol.83, No.1, 1996, pp.167-188.

[4] Dimou, C. and Koumousis, V., "Genetic algorithms in a competitive environment with application to reliability optimal design", *Proceedings of the sixth international conference on Application of artificial intelligence to civil & structural engineering*, 2001, pp.89-90.

[5] Goldberg, D.E. and Deb, K., "A Comparative Analysis of Selection Schemes Used in Genetic Algorithms", *Foundations of Genetic Algorithms*, 1991, pp.69-93.

[6] Hartl, D.L. and Clark, A.G., "Principles of Population Genetics", Sinauer Associates, 1997

[7] Hemaspaandra, L.A. and Ogihara, M., "The Complexity Theory Companion", Springer-Verlag, 2001

[8] Holland, J.H., "Genetic algorithms and the optimal allocation of trials", *SIAM Journal on Computing*, Vol.2, 1973, pp.88-105.

[9] Holland, J.H., "Adaptation in Natural and Artificial Systems", University of Michigan Press, 1975

[10] Julstrom, B.A., "Seeding the population: improved performance in a genetic algorithm for the rectilinear Steiner problem", *Proceedings of the 1994 ACM symposium on Applied computing*, 1994, pp.222-226.

[11] Koza, J.R., "Genetic Programming: On the Programming of Computers by Means of Natural Selection", MIT Press, 1992

[12] Mahfoud, S. and Mani, G., "Financial forecasting using genetic algorithms", *Applied Artificial Intelligence*, Vol.10, No.6, 1996, pp.543-565.

[13] Mehta, A., Saberi, A., Vazirani, U., Vazirani, V. and Mehta, A., "AdWords and Generalized On-line Matching", *Proceedings of 46th Annual IEEE Symposium on Foundations of Computer Science*, 2005, pp.264-273.

[14] Michalewicz, Z., "Genetic Algorithms + Data Structures = Evolution Programs", Springer, 1998

[15] Miller, B.L. and Goldberg, D.E., "Genetic Algorithms, Selection Schemes and the Varying Effects of Noise", *Evolutionary Computation*, Vol.4, 1996, pp.113-131.

[16] Raghuwanshi, M.M. and Kakde, O.G., "Survey on multiobjective evolutionary and real coded genetic Algorithms", *Proceeding of the 8th Asia Pacific Symposium on Intelligent and Evolutionary Systems*, 2004, pp.150-161.

[17] Sambridge, M. and Gallagher, K., "Earthquake hypocenter location using genetic algorithms", *Bulletin of the Seismological Society of America*, Vol.83, No.5, 1993, pp.1467-1491.

[18] Sato, S., Otori, K., Takizawa, A., Sakai, H., Ando, Y. and Kawamura, H., "Applying genetic algorithms to the optimum design of a concert hall", *Journal of Sound and Vibration*, Vol.258, No.3, 2002, pp.517-526.

[19] Schirmer, A., "A Guide to Complexity Theory in Operations Research", *Manuskripte aus den Instituten fur Betriebswirtschaftslehre der Universitat Kiel*, 1995

[20] Schwefel, H-P., "Special Track On Computational Intelligence – Genetic Algorithms", *Proceedings of the 23rd Euromicro Conference*, 1997, pp.622-623.

[21] Sliwko, L. and Zgrzywa, A., "Multi-resource load optimization strategy in agent-based systems", *Lecture Notes in Artificial Intelligence 4496*, 2007, pp.348–357.

[22] Yagiura, M. and Ibaraki, T. , "On metaheuristic algorithms for combinatorial optimization problems", *Systems and Computers in Japan*, Vol.32, No.3, 2001, pp.33-55.